\documentclass{article}
\usepackage{ijcai21}
\usepackage{times}
\usepackage{soul}
\usepackage{url}
\usepackage[hidelinks]{hyperref}
\usepackage[utf8]{inputenc}
\usepackage[small]{caption}
\usepackage{graphicx}
\usepackage{amsmath}
\usepackage{amsthm}
\usepackage{booktabs}
\usepackage{algorithm}
\usepackage{gensymb}
\usepackage{algorithmic}
\usepackage{xcolor}
\usepackage{comment}
\usepackage{multirow}
\title{WildGait: Learning Gait Representations from Raw Surveillance Streams}
\author{
Adrian Cosma\footnote{Corresponding Author}$^1$
\and
Emilian Radoi$^1$\and
\affiliations
$^1$University Politehnica of Bucharest\\
\emails
cosma.i.adrian@gmail.com,
emilian.radoi@cs.pub.ro
}

\begin{document}

\maketitle

\begin{abstract}
The use of gait for person identification has important advantages such as being non-invasive, unobtrusive, not requiring cooperation and being less likely to be obscured compared to other biometrics. Existing methods for gait recognition require cooperative gait scenarios, in which a single person is walking multiple times in a straight line in front of a camera. We aim to address the challenges of real-world scenarios in which camera feeds capture multiple people, who in most cases pass in front of the camera only once. We address privacy concerns by using only motion information of walking individuals, with no identifiable appearance-based information. As such, we propose a novel weakly supervised learning framework, WildGait, which consists of training a Spatio-Temporal Graph Convolutional Network on a large number of automatically annotated skeleton sequences obtained from raw, real-world, surveillance streams to learn useful gait signatures. We collected the training data and compiled the largest dataset of walking skeletons called Uncooperative Wild Gait, containing over 38k tracklets of anonymized walking 2D skeletons. We release the dataset for public use.
Our results show that, with fine-tuning, we surpass the current state-of-the-art pose-based gait recognition solutions. Our proposed method is reliable in training gait recognition methods in unconstrained environments, especially in settings with scarce amounts of annotated data. 

\end{abstract}

\section{Introduction}
\begin{figure}[hbt!]
    \centering
    \includegraphics[width=\linewidth]{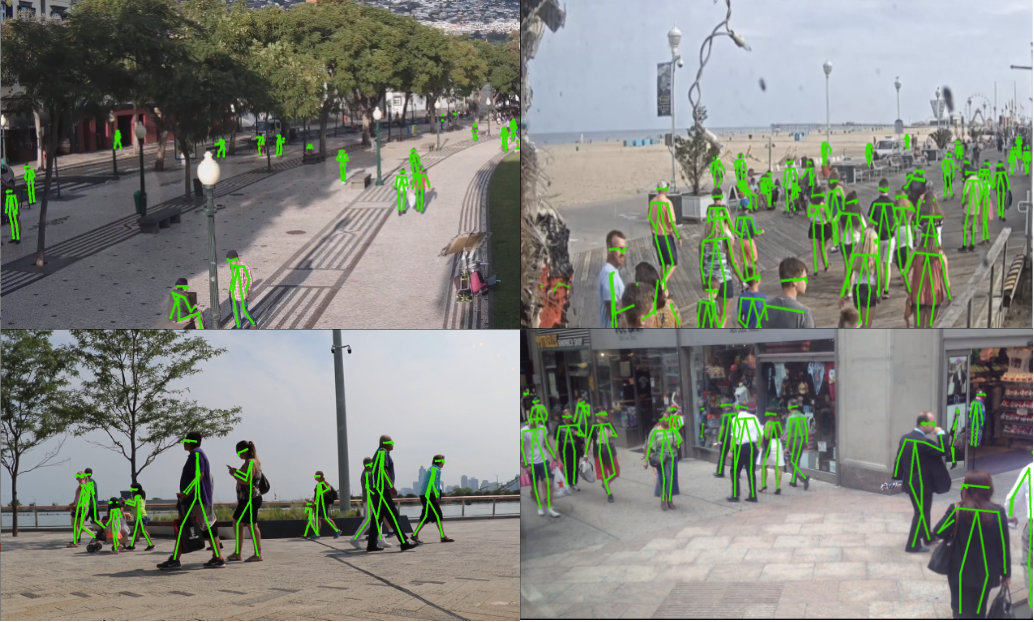}
    \caption{Samples from the UWG dataset. We gather skeleton sequences by applying state-of-the-art pose estimation / pose tracking models on publicly available surveillance streams. This allows for learning discriminative gait representations without explicit human labels and without the cooperation of subjects.}
    \label{fig:samples}
\end{figure}

% \begin{figure*}[hbt!]
%     \centering
%     \includegraphics[width=0.75\textwidth]{images/WildGait3-batchnorm.png}
%     \vspace{-2mm}
%     \caption{The WildGait framework. We process raw video streams from real-world settings of people walking and construct Uncooperative Wild Gait (UWG), a dataset of skeleton sequences loosely annotated through pose estimation and pose tracking. We use this raw and noisy data to pretrain a gait recognition model that generalizes well to different gait recognition scenarios.}
%     \label{fig:framework}
% \end{figure*}

The study of gait (manner of walking) has gained increased attention in recent years, as it encodes important behavioral biometric information, and the recent advancements in machine and deep learning provide the necessary toolset to model this information. Walking patterns can be used to estimate the age and gender of a person \cite{10.1007/978-981-15-5148-2_82}, estimate emotions \cite{randhavane2020identifying}, and provide insight into various physiological conditions \cite{gaitbook}. Moreover, aside from these soft-biometrics, gait information is used as a unique fingerprinting method for identifying individuals. Although face recognition has become the norm for person identification in a range of suitable applications with good results, gait recognition from video is still a challenging task in real-world scenarios. The intrinsic dynamic nature of walking makes it susceptible to a multitude of confounding factors such as view angle, shoes and clothing, carrying variations, age, interactions with other people and various actions that the person is performing while walking. 

This led the study of gait recognition to be mostly performed in controlled environments, in which only a subset of confounding factors are explored \cite{1699873}. Current datasets \cite{1699873,An_TBIOM_OUMVLP_Pose,gait-recognition-via-disentangled-representation-learning} focus on the change in view angle, clothing and carrying conditions, while ignoring other behavioral variations. 
As opposed to face recognition datasets, gait recognition datasets are harder to build, and require the cooperation of thousands of subjects in order to be relevant. Furthermore, privacy laws restrict the usage of these datasets in real world applications.

We propose WildGait, a framework for automatically learning useful, discriminative representations for gait recognition from raw, real-world data, without explicit human labels. We leverage surveillance streams of people walking and employ state of the art pose estimation methods (e.g. AlphaPose \cite{li2018crowdpose}) to extract skeleton sequences and pose tracking to construct a loosely annotated dataset. Making use of a diverse set of augmentation procedures, we pretrain our network with minimal direct supervision - the only indirect label employed is the pose tracking information uncovered automatically.

We chose to use pose estimations as they do not contain any identifiable appearance-based information of walking individuals. Current methods that use silhouettes \cite{1561189} are unsuitable in real-world, dynamic scenarios in which changes in illumination and multiple overlapping individuals severely affect the quality of extracted silhouettes. While some approaches explicitly disentangle appearance features and pose information \cite{gait-recognition-via-disentangled-representation-learning}, we argue that appearance-based methods do not respect the privacy of individuals. Skeletons extracted from human pose estimation methods encode only motion information, which is sufficient to determine if two skeleton representations belong to the same person, without holding any information about the person's identity. 
Pose information also enables leveraging information of performed actions and activities, and filtering out individuals that are not walking or have abnormal walking patterns \cite{8653351}.

For pretraining, we propose the Uncooperative Wild Gait dataset (UWG), which unlike current available datasets, contains anonymised skeleton sequences of a large number of people walking in a natural environment (over 38k tracklets), with many walking variations and confounding factors - the data is gathered from raw, real-world video streams (Figure \ref{fig:samples}). People walking in UWG are present only once, from a single viewing angle and with a constant array of intrinsic confounding factors (clothing, shoes etc.), making it a highly challenging dataset. We leverage this noisy information to pretrain a neural network to better handle controlled gait sequences in scenarios with few data samples.

This paper makes the following contributions:

\begin{enumerate}
    \item  We are among the first to explore unsupervised learning on gait recognition, and propose a novel framework, WildGait, which describes a data collection and pretraining pipeline, that enables learning meaningful gait representations in a weakly supervised manner, from automatically extracted skeleton sequences in unconstrained environments. 
    
    \item The Uncooperative Wild Gait dataset (UWG), dataset of noisily tracked skeleton sequences to allow the gait recognition research community to further explore ways to pretrain gait recognition systems in an unsupervised manner.
    
    \item A study on transfer learning capabilities of our pretrained network on popular gait recognition databases, highlighting great data-efficiency when fine-tuning.
    
\end{enumerate}

\section{Related Work}
\begin{figure*}[hbt!]
    \centering
    \includegraphics[width=\textwidth]{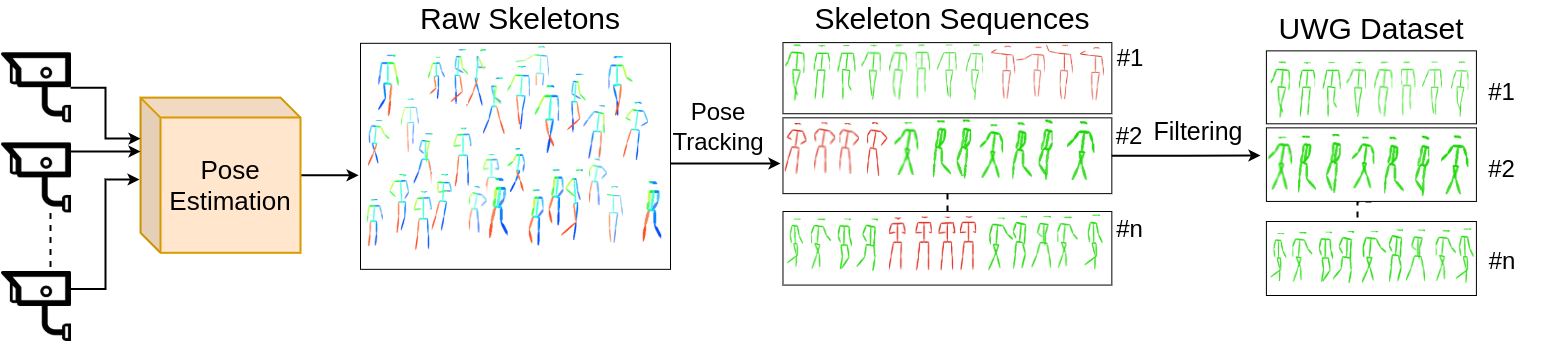}
    \vspace{-2mm}
    \caption{Data collection procedure. We process raw video streams from real-world settings of people walking and construct Uncooperative Wild Gait (UWG), a dataset of skeleton sequences loosely annotated through pose estimation and pose tracking. We use this raw and noisy data to pretrain a gait recognition model that generalizes well to different gait recognition scenarios.}
    \label{fig:framework}
\end{figure*}

% gait recognition: GEI stuff
The research attention received by gait recognition over the past decade has been increasing. A significant portion of this research was dedicated to gait recognition using wearable inertial sensors \cite{Sprager2015}, however, our focus is on recognition of gait using camera sensors.  Gait recognition approaches can be classified into two main categories: model-based and appearance-based. One of the most prevalent approaches for appearance-based gait recognition is the use of a Gait Energy Image (GEI) \cite{1561189}. GEIs are computed by averaging the silhouettes of walking individuals across a gait cycle. Such images are then processed using modern standard image processing approaches. Since GEIs have limitations, such as not taking into account temporal information, several variants are proposed to address these shortcomings, notably Gait Entropy Image (GEnI) \cite{gait-entropy-image}, Gait Flow Image \cite{LAM2011973} and Chrono-Gait Image \cite{10.1007/978-3-642-15549-9_19}, all showing good performance on benchmark datasets. 

More recent approaches tend to use appearance features to explicitly learn a disentangled representation. \cite{gait-recognition-via-disentangled-representation-learning} propose a way to explicitly disentangle motion and appearance features via an autoencoder with carefully designed loss functions.

% Several benchmark datasets have been proposed to test the performance of gait recognition systems in the presence of a standard array of confounding factors. The most popular dataset is CASIA-B \cite{1699873}, which is comprised of 126 different identities, that walk several times in front of 11 cameras. Each identity also has different walking variations (clothing / carrying conditions) that affect the walking patterns. More recently, Front-View Gait Database (FVG) \cite{gait-recognition-via-disentangled-representation-learning} was proposed to tackle the most difficult camera view-angle for gait processing systems (i.e. walking towards the camera). The authors propose several confounding factors, including clothing and carrying conditions, background clutter, and the passage of time. Another recent dataset, OU-MVLP \cite{An_TBIOM_OUMVLP_Pose}, is introduced as a benchmark for evaluating the scalability of gait representations, being one of the largest datasets, comprised of 10.000 identities captured from 14 cameras.

As opposed to appearance-based methods, model-based methods process walking patterns as a set of human joint trajectories across time. The performance increase of pose-estimation models \cite{Chen_2020}, enabled the use of skeletons sequences in gait recognition. \cite{7899654} directly use the joint probability heatmaps as an input for an LSTM network. \cite{ptsn} used an LSTM and a CNN to process 2D skeleton sequences to account for the temporal and spatial variations of walking. \cite{SHENG202086} used an LSTM autoencoder with contrastive learning to further stabilize the joint trajectories of skeletons. Recently, \cite{8653351} enhance the robustness of estimated skeletons by constructing a quality-adjusted cost matrix between input frames and registered frames for frame-level matching.  \cite{Lima2020} propose a a fully-connected network to model a single skeleton, and temporal aggregation of skeletal features for the final classification. Both \cite{LIAO2020107069} and \cite{10.1007/978-3-319-97909-0_15} leverage 3D skeletons to model gait patterns, with incremental improvements over 2D skeletons. Similar to us, \cite{li2020jointsgaita} applies a graph convolutional network to process skeleton sequences, but use a final pyramid pooling layer for recognition. An important aspect of gait recognition for deployment in practice is  multi-gait, in which multiple people walk together, and their individual walking patterns change. \cite{7976333} proposed an attribute discovery model in a max-margin framework to recognize a person based on gait while walking with multiple people.

Little research was performed on gait recognition in more constrained scenarios, with little to no annotated data. Closer to the work proposed in this paper are the recent advancements in skeleton-based action recognition in scenarios of little or no supervision. \cite{su2019predict} proposed to use an LSTM autoencoder to learn discriminative representations for activity recognition, without any supervision except the skeleton sequences. Similarly, \cite{li2020iterate} used an iterative approach in an active learning setting. \cite{Lin_2020} used a self-supervised approach to activity recognition, in which they propose several pretext tasks to pretrain the network, such as pose shuffling and motion prediction. However, different from the previous action recognition methods which aim to ambiguate the identity from various actions, we target the opposite problem: given a single action (i.e. walking), we aim to uncover the identity. As such, we posit that directly using unsupervised action recognition methods for pretraining is unsuitable for our case.

Different from self-supervised methods, we take a data-driven approach. We process large amounts of raw data from real-world video streams, automatically extracting skeletons from each frame, performing intra-frame pose tracking and filtering unwanted skeletons (i.e. poorly extracted / non-walking). We train an ST-GCN using contrastive learning for gait recognition without any major architectural modifications, and obtain exceptional results in scenarios with scarce amounts of data.

\section{Method}
\subsection{Dataset Construction}

Our aim is to learn good gait representations from human skeleton sequences in unconstrained environments, without explicit labels. For this purpose we collect a sizeable dataset of human walking skeleton sequences from surveillance camera feeds, with a high variance of walking styles and from various geographic locations, environments, weather conditions and camera angles. The dataset captures a multitude of confounding factors in the manner of walking of individuals. We named this dataset Uncooperative Wild Gait (UWG), and publicly released it for the research community to further advance the field of gait recognition.

\begin{figure}[hbt!]
    \centering
    \includegraphics[width=0.75\linewidth]{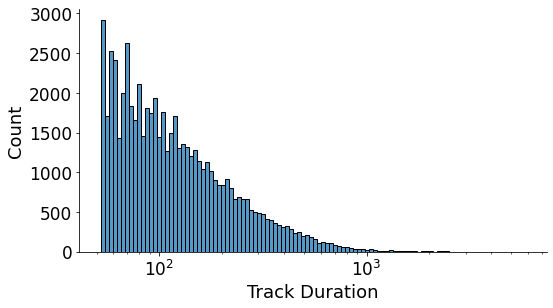}
    % \vspace{-2mm}
    \caption{Distribution of track durations in the UWG dataset.}
    \label{fig:track-duration}
\end{figure}

Since people from different geographic locations have different manners of walking influenced by cultural and societal norms \cite{AlObaidi2003}, we gathered data uniformly from 3 major continents (Europe, Asia and North America - Table \ref{tab:continents}).
To obtain skeleton sequences for people in a video, we first preprocess the streams to a common 24fps and 720p resolution. Then, a crowd pose estimation method (e.g. AlphaPose \cite{li2018crowdpose}) was used to extract human skeletons for each frame, and skeletons were tracked intra-camera, across time, with a simple Kalman filter \cite{Bewley_2016}. The output of the AlphaPose model consists of 17 joints with X and Y coordinates of the joints and the confidence, in the COCO \cite{DBLP:journals/corr/LinMBHPRDZ14} pose format.

\setlength{\tabcolsep}{0.6em}
{\renewcommand{\arraystretch}{1.2}
    \begin{table}[hbt!]
        \small
        \centering
    \resizebox{1.0\linewidth}{!}{
        \begin{tabular}{r || r | r |r}
            \textbf{Continent} & \textbf{\# IDs$^*$} & \textbf{Walk Length (hr)} & \textbf{Avg. Run Len. (frames)} \\
                 \hline\hline
                
            Asia & 3635 & 4.79 & 104.5 \\
            Europe & 12993 & 19.76 & 110.1 \\
            North America & 21874 & 34.47 & 108.2 \\
        \end{tabular}
    }
        \caption{Geographic locations from where we gathered skeleton sequences. We aimed for a multi-cultural representation of walking people. $^*$ Approximate number given by pose tracker.} 
        \label{tab:continents}
    \end{table}
}

Unstructured environments invariably introduce noise in the final skeleton sequences due to unreliable extracted skeletons, lost tracking information, or people standing who perform other activities than walking. To address this, we only keep sequences with a mean confidence on extracted joints of over 60\% and with no more than 3 consecutive frames with feet confidence of less than 50\%. This way we eliminate poorly extracted poses and ensure that the feet are visible throughout the sequence. Moreover, we enforce a minimum tracking sequence of 54 frames, which corresponds to approximately two full gait cycles \cite{00004623-196446020-00009}.
Further, based on known body proportions we normalize each skeleton to be invariant to the height of the person by first zero-centering each skeleton by subtracting the pelvis coordinates, and then normalizing the Y coordinate by the length between the neck and the pelvis, and the X coordinate by length from the right to the left shoulder (Equations \ref{eq:eq1} and \ref{eq:eq2}). The normalization procedure ensures the skeleton sequences are aligned and similar poses have close coordinates. Moreover, by removing information related to the height of the person, information related to the stature and particular body characteristics of a person are eliminated. 

\begin{equation}
    x_{joint} = \frac{x_{joint} - x_{pelvis}}{|x_{R.shoulder} - x_{L.shoulder}|}
    \label{eq:eq1}
\end{equation}
\begin{equation}
    y_{joint} = \frac{y_{joint} - y_{pelvis}}{|y_{neck} - y_{pelvis}|}
    \label{eq:eq2}
\end{equation}

We address the issue of non-walking people with an heuristic on the movement of the joints corresponding to the legs (feet and knees). We compute the average movement velocity of the feet, which is indicative of the activity of a person. Thus, we filter out individuals with an average velocity magnitude of less than 0.01. Extremely long tracklets are also filtered out, as it was noticed that individuals tracked for a longer time are usually standing (not walking). 

The proposed framework does not rely on appearance information at any step in the processing pipeline, except when extracting the pose information. A total of 38.502 identities were obtained, with an average walking duration of 108 frames. The total walking sequences duration in the dataset is of approximately 60 hours. The scenario for this dataset is more restrictive compared to other benchmark datasets, as it does not include multiple runs of the same person from multiple camera angles and with multiple walking styles (such as different carrying and clothing conditions). Still, even from this restrictive scenario, the large amount of data is leveraged to learn good gait representations that transfer well to other benchmark datasets. In the proposed configuration, the UWG dataset is used only for training, and the learned embeddings are evaluated on popular gait recognition datasets.

\setlength{\tabcolsep}{0.6em}
{\renewcommand{\arraystretch}{1.2}
    \begin{table}[h]
        \small
        \centering
    \resizebox{\linewidth}{!}{
        \begin{tabular}{p{0.2\linewidth} || p{0.1\linewidth} | p{0.08\linewidth} | p{0.18\linewidth} | p{0.15\linewidth} | p{0.1\linewidth}}
        \textbf{Dataset} & \textbf{\# IDs} & \textbf{Views} & \textbf{Total Walk Length (hr)} & \textbf{Avg. Run Length (frames)} & \textbf{Runs / ID} \\
             \hline\hline
        CASIA-B & 124 & 11 & 15.8 & 100 & 110 \\
        \hline
        FVG & 226 & 3 & 3.2 & 97 & 12 \\
        \hline\hline
         UWG (\textbf{ours)} & \textbf{38.502$^*$} & \textbf{1} & \textbf{59.0} & \textbf{108.5} & \textbf{1}\\
        \end{tabular}
    }
         \vspace{-2mm}
        \caption{Comparison of gait datasets. UWG differs in purpuse, as it is intended for pretraining, and not for evaluation. It is a large-scale dataset, noisily annotated, agnostic to confounding factors and camera viewpoints. $^*$ Approximate number given by pose tracker.} 
        \label{tab:stats}
    \end{table}
}

\subsection{Learning Procedure}

\begin{figure*}[hbt!]
    \centering
    \includegraphics[width=0.75\linewidth]{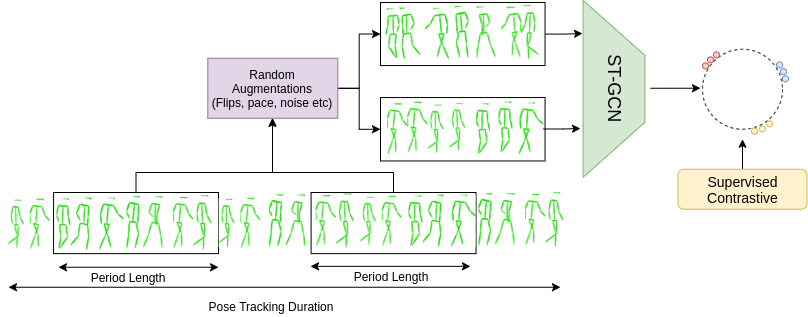}
    \vspace{-2mm}
    \caption{WildGait training methodology. Given a tracked sequence of skeletons, we employ Supervised Contrastive (SupCons) loss on multiple augmented views (randomly cropped, flipped etc) of the same person.}
    \label{fig:training}
\end{figure*}

Figure \ref{fig:training} highlights the proposed methodology for learning informative gait representations from unconstrained scenarios. Walking sequences for each tracked person are obtained after processing the data from the video streams. A Spatio-Temporal Graph Convolutional Network (ST-GCN) \cite{yan2018spatial} is employed to process the walking sequences, which was chosen due to its good results in the area of action recognition. Moreover, applying graph computation on skeletons allows modelling both the local interactions between joints and the global time variation between individual skeletons. A graph model was used for the implementation, as it is more appropriate to model the spatio-temporal relationships between joints compared to a simple LSTM network \cite{7899654}.  Moreover, \cite{li2020jointsgaita} show that a ST-GCN can successfully be used for skeleton-based gait recognition.

Since the setting for the UWG dataset does not include multiple runs of the same walking person, with different confounding factors, a diverse set of data augmentations are employed to create augmented walking sequences for the same person. The network receives a randomly sampled gait sequence of 54 frames, out of the full tracked sequence. If the tracked sequence is less than 54 frames, the start of the sequence is repeated. Moreover, the sequence is dilated or contracted, in accordance to pace prediction \cite{wang2020selfsupervised}. Modifying the pace of a video sequence has been shown to allow for learning meaningful semantic information in a self-supervised manner. The time modification factor is uniformly sampled from $\{0.25, 0.5, 0.75, 1, 1.25, 1.5, 1.75, 2\}$. This procedure allows for the model to be robust to changes in video framerate and subject walking speed.

Further, portions of the skeleton sequence are randomly shuffled to model temporal patterns, similarly to \cite{Lin_2020}. Shuffling the order of the skeleton sequence introduces spatio-temporal ambiguity, that forces to model to learn invariant representations for use in downstream tasks.

Finally, squeezing, flipping, mirroring and randomly dropping out joints and frames are employed to further introduce diversity in the dataset.

Existing literature training procedures for recognition problems, employ either triplet loss \cite{DBLP:journals/corr/SchroffKP15}, center loss \cite{10.1007/978-3-319-46478-7_31}, direct classification, or a combination of them \cite{luo2019bag}. However, manipulating the loss weights of multiple loss functions is cumbersome, are requires significant tuning. Moreover, having a direct classification head of the person identity does not scale well with the number of identities, and many models employ such a classification head for regularisation and preventing the triplet loss from collapsing.

As such, we employed the Supervised Contrastive (SupCons) 
\cite{khosla2020supervised} loss, which is a generalization of the triplet loss, allowing for multiple positive examples per identity. We chose two randomly augmented views of the same identity for training. Figure \ref{fig:training} showcases our simplified learning procedure. After the feature extraction, embeddings are normalized to the unit sphere. Following the author's recommendations, we employ a loss temperature of 0.01, as smaller temperatures benefits training.

By not using a direct classification head of the identities, as in previous gait recognition works, we are able to scale to tens of thousands of identities with the same model size. This was not a requirement when working with smaller datasets such as CASIA-B and FVG, but it is in the case of UWG, since it has 38k identities.

\section{Experiments and Results}
Popular gait recognition datasets, CASIA-B and FVG (Front-View Gait), were chosen to evaluate our unsupervised pretraining scheme. We chose CASIA-B as it is a popular benchmark for evaluating the effect of viewpoint variation, which allows us to compare to other pose-based gait recognition methods. FVG is used to further evaluate the front-view angle, across multiple confounding factors (i.e. carrying conditions / clothing change / passage of time etc).

The performance evaluations presented in this paper abide by the evaluation guidelines of each dataset. For CASIA-B, we use the first 62 identities for training and the final 62 for evaluation, and show the average recognition accuracy across all viewing angles, except when gallery and probe angles are the same. For FVG, we used 136 identities for training and the rest for testing, and report results for each evaluation protocol: Walk Speed (WS), Carrying Bag (CB), Changing Clothes (CL), Cluttered Background (CBG) and ALL. 

\begin{figure}[hbt!]
    \centering
    \includegraphics[width=1.0\linewidth]{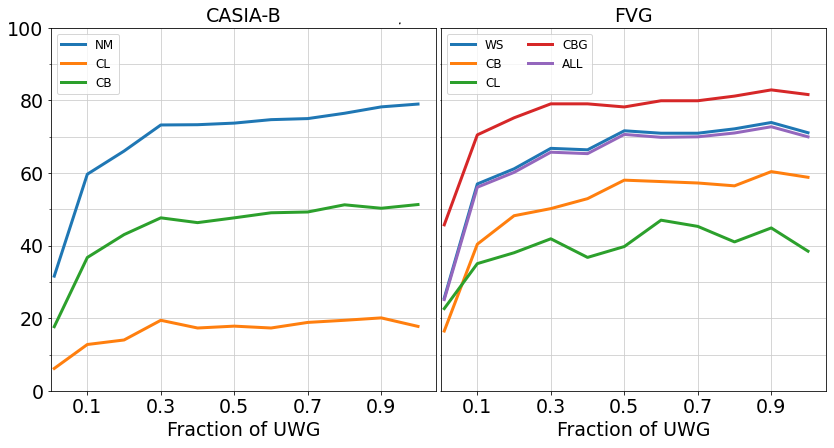}
    \caption{UWG dataset size influence over the transfer learning performance of WildGait on downstream evaluation benchmarks. The network was pretrained on UWG, but not trained on CASIA-B or FVG. For CASIA-B, we show mean accuracy where the gallery set contains all viewpoints except the probe angle.}
    \label{fig:dataset-size}
\end{figure}

We initially tested our network's capability to generalize to CASIA-B and FVG without actually training or fine-tuning on these datasets. We evaluated the transfer recognition accuracy using increasingly larger random samples from UWG to highlight the impact of the size of the pretraining dataset. Each experiment was run 5 times and the results were averaged to avoid a favorable configuration for our setting. The results in Figure \ref{fig:dataset-size} show that transfer accuracy on downstream tasks benefits from a larger size of the pretraining dataset. It is more evident in the case of FVG, since it has fewer viewpoint variations pertaining to each subject.

\begin{figure}[hbt!]
    \centering
    \includegraphics[width=1.0\linewidth]{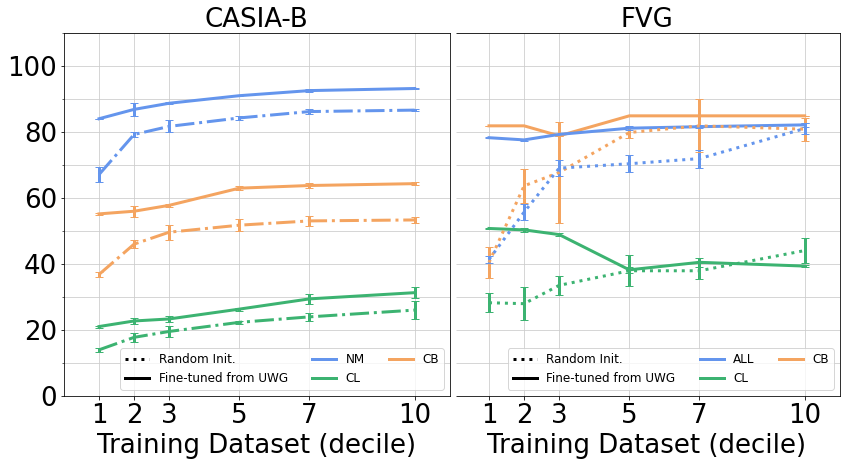}
    \caption{Performance of fine-tuning the proposed network on downstream evaluation benchmarks, with fractions of the training data. For FVG, runs are randomly sampled per subject. For CASIA-B, runs are randomly sampled uniformly from all angles per subject, and the average accuracy is across all viewpoints, where the gallery contains all angles except the probe angle. Pretraining on UWG results in a more stable training regime and significantly increased performance in scenarios with little labelled training data available.}
    \label{fig:fine-tune}
\end{figure}

Further, we evaluated the performance of our pretrained network when fine-tuned using limited amounts of training data. For both CASIA-B and FVG we used random samples of 10\%, 20\%, 30\%, 50\%, 70\% and 100\% of the runs of each person. Each experiment was run 5 times and the results were averaged. The entire network was trained with a high learning rate at deeper levels in the architecture and a decreasingly lower rate for the lower-level representations, as proposed by \cite{DBLP:journals/corr/KirkpatrickPRVD16}.
The pretrained network was compared to a randomly initialized one, with results presented in Figure \ref{fig:fine-tune}. 
On both datasets, the benefits of pretraining are showcased in the constant superior performance over random weight initialization, especially with lower amounts of training data. Moreover, the training regime is more stable regardless of the amount of data.

\setlength{\tabcolsep}{0.8em}
{\renewcommand{\arraystretch}{1.2}
\begin{table*}[hbt!]
    \resizebox{\textwidth}{!}{
    \centering
    \begin{tabular}{l | ccccccccccc | c }
    \toprule
        & \multicolumn{12}{ c| }{\textbf{CASIA-B - Normal Walking}}\\
         &  0$^{\circ}$  &    18$^{\circ}$  &    36$^{\circ}$  &    54$^{\circ}$  &    72$^{\circ}$  &    90$^{\circ}$  &    108$^{\circ}$ &    126$^{\circ}$ &   144$^{\circ}$ &    162$^{\circ}$ &    180$^{\circ}$ & Mean \\
        \midrule
        Pretrained Kinetics  & 19.35 &  19.35 &  27.42 &  29.03 &  25.81 &  38.71 &  27.42 &  27.42 &  20.97 &  12.9 &  6.45 &  23.17 \\
        \hline
        Predict \& Cluster & 41.93 &  45.16 &  45.96 &  34.67 &  20.16 &  11.29 &  30.64 &  32.25 &  21.77 &  19.35 &  12.09 &  28.66 \\
        \hline
        MS$^2$L & 37.90 &  39.51 &  40.32 &  51.61 &  24.19 &  17.74 &  25.80 &  46.74 &  40.32 &  33.06 &  34.67 &  35.62 \\
        Pace Prediction  & 43.34 &  39.51 &  50.00 &  54.03 &  38.70 &  62.90 &  \textbf{70.96} & 70.16 &  54.03 &  66.93 &  64.51 &  55.91\\
        \hline
        WildGait \textbf{(ours)} & \textbf{72.58} &  \textbf{84.67} &  \textbf{90.32} &  \textbf{83.87} &  \textbf{63.70} &  \textbf{62.90} &  66.12 &  \textbf{83.06} &  \textbf{86.29} &  \textbf{84.67} &  \textbf{83.06} &  \textbf{78.29} \\
    \end{tabular}}
    \vspace{-2mm}
    \caption{Transfer learning comparison with other unsupervised skeleton-based training methods on CASIA-B. We report accuracy where the gallery set contains all viewpoints except the probe angle (top row).} 
    \label{tab:unsupervised-casia}
\end{table*}
}
\setlength{\tabcolsep}{0.8em}
{\renewcommand{\arraystretch}{1.3}
\begin{table}[hbt!]
    \centering
    \resizebox{1.0\linewidth}{!}{
    \begin{tabular}{l | ccccc}
    \toprule
        & \multicolumn{5}{ c }{\textbf{FVG}} \\
         & WS & CB & CL & CBG & ALL \\
        \midrule
        Pretrained Kinetics & 24.00 & 54.55 & 28.63 & 43.16 & 22.33 \\
        \hline
        Predict \& Cluster &  32.79 &  33.72 &  20.08 &  44.01 &  32.40 \\
        \hline
        MS$^2$L &  42.33 &  40.78 &  31.62 &  53.84 &  41.93 \\
        Pace Prediction & 45.65 &  40.78 &  28.63 &  55.55 &  44.84 \\
        \hline
        WildGait \textbf{(ours)} &  \textbf{75.66} & \textbf{81.81} & \textbf{48.71} & \textbf{84.61} & \textbf{75.66} \\
    \end{tabular}
    }
    \vspace{-2mm}
    \caption{Transfer learning comparison with other unsupervised skeleton-based training methods on FVG dataset.} 
    \label{tab:unsupervised-fvg}
\end{table}
}
\setlength{\tabcolsep}{0.7em}
{\renewcommand{\arraystretch}{1.02}
\begin{table*}[hbt!]
    \resizebox{1.0\textwidth}{!}{
    \centering
    \begin{tabular}{l l | rrrrrrrrrrr| r}
    \toprule
        Probe & Method &  0$^{\circ}$  &    18$^{\circ}$  &    36$^{\circ}$  &    54$^{\circ}$  &    72$^{\circ}$ & 90$^{\circ}$  &  108$^{\circ}$ &    126$^{\circ}$ &   144$^{\circ}$ &    162$^{\circ}$ & 180$^{\circ}$ & Mean \\
        \midrule
        \multirow{6}{*}{NM} & PTSN \cite{ptsn} & 34.5 & 45.6 & 49.6 & 51.3 & 52.7 & 52.3 & 53 & 50.8 & 52.2 & 48.3 & 31.4 & 47.4 \\
        & PTSN-3D \cite{10.1007/978-3-319-97909-0_15} & 38.7 & 50.2 & 55.9 & 56 & 56.7 & 54.6 & 54.8 & 56 & 54.1 & 52.4 & 40.2 & 51.9 \\
        & PoseGait \cite{LIAO2020107069} & 48.5 & 62.7 & 66.6 & 66.2 & 61.9 & 59.8 & 63.6 & 65.7 & 66 & 58 & 46.5 & 60.5 \\
        % & JointsGait \cite{li2020jointsgaita} & 68.1 & 73.6 & 77.9 & 76.4 & 77.5 & 79.1 & 78.4 & 76 & 69.5 & 71.9 & 70.1 & 74.4 \\
        & PoseFrame \cite{Lima2020} & 66.9 & 90.3 & 91.1 & 55.6 & 89.5 & \textbf{97.6} & \textbf{98.4} & 97.6 & 89.5 & 69.4 & 68.5 & 83.1 \\
        & WildGait network \textbf{(ours)} & \textbf{86.3} &    \textbf{96.0} &    \textbf{97.6} &    \textbf{94.3} &    \textbf{92.7} &    94.3 &     94.3 &     \textbf{98.4} & \textbf{97.6} &     \textbf{91.1} & \textbf{83.8} & \textbf{93.4} \\
 
        \hline
        \multirow{6}{*}{CL} & PTSN \cite{ptsn} & 14.2 & 17.1 & 17.6 & 19.3 & 19.5 & 20 & 20.1 & 17.3 & 16.5 & 18.1 & 14 & 17.6 \\
        & PTSN-3D \cite{10.1007/978-3-319-97909-0_15} & 15.8 & 17.2 & 19.9 & 20 & 22.3 & 24.3 & 28.1 & 23.8 & 20.9 & 23 & 17 & 21.1 \\
        & PoseGait \cite{LIAO2020107069} & 21.3 & 28.2 & 34.7 & 33.8 & 33.8 & 34.9 & 31 & 31 & \textbf{32.7} & 26.3 & 19.7 & 29.8 \\
        % & JointsGait \cite{li2020jointsgaita} & \textbf{48.1} & \textbf{46.9} & \textbf{49.6} & \textbf{50.5} & \textbf{51} & 52.3 & 49 & 46 & \textbf{48.7} & \textbf{53.6} & \textbf{52} & \textbf{49.8} \\
        & PoseFrame \cite{Lima2020} & 13.7 & 29.0 & 20.2 & 19.4 & \textbf{28.2} & \textbf{53.2} & \textbf{57.3} &\textbf{52.4} & 25.8 & 26.6 & 21.0 & 31.5 \\
        & WildGait network \textbf{ (ours)} & \textbf{29.0} &    \textbf{32.2} &    \textbf{35.5} &    \textbf{40.3} &    26.6 &    25.0 &     38.7 &     38.7 &     31.4 &     \textbf{34.6} &     \textbf{31.4} & \textbf{33.0}\\

        \hline
        \multirow{6}{*}{BG} & PTSN \cite{ptsn} & 22.4 & 29.8 & 29.6 & 29.2 & 32.5 & 31.5 & 32.1 & 31 & 27.3 & 28.1 & 18.2 & 28.3 \\
        & PTSN-3D \cite{10.1007/978-3-319-97909-0_15} & 27.7 & 32.7 & 37.4 & 35 & 37.1 & 37.5 & 37.7 & 36.9 & 33.8 & 31.8 & 27 & 34.1 \\
        & PoseGait \cite{LIAO2020107069} & 29.1 & 39.8 & 46.5 & 46.8 & 42.7 & 42.2 & 42.7 & 42.2 & 42.3 & 35.2 & 26.7 & 39.6 \\
        % & JointsGait \cite{li2020jointsgaita} & 54.3 & 59.1 & 60.6 & 59.7 & 63 & 65.7 & 62.4 & 59 & 58.1 & 58.6 & 50.1 & 59.1 \\
        & PoseFrame \cite{Lima2020} & 45.2 & 66.1 & 60.5 & 42.7 & \textbf{58.1} & \textbf{84.7} & \textbf{79.8} & \textbf{82.3} & \textbf{65.3} & 54.0 & 50.0 & 62.6 \\
        & WildGait network \textbf{(ours)} & \textbf{66.1} &    \textbf{70.1} &    \textbf{72.6} &    \textbf{65.3} &    56.4 &    64.5 &     65.3 &     67.7 &     57.2 &     \textbf{66.1} &     \textbf{52.4} & \textbf{64.0} \\
    \end{tabular}}
    \vspace{-2mm}
    
    \caption{Comparison with other skeleton-based gait recognition methods on CASIA-B dataset. The evaluation was performed in the "leave-one-out" setting, in which the gallery set contains all viewpoints except the one in the probe set. WildGait achieves state-of-the art results in normal walking (NM) and carry-bag variation (CB) by a large margin, being able to generalize well across camera viewpoints. }
    \label{tab:casia-sota}
\end{table*}
}

We compared WildGait to other relevant methods that leverage skeleton sequences to learn meaningful representations. 
A ST-GCN pretrained on Kinetics \cite{kay2017kinetics} was selected to evaluate the transfer learning capabilities from supervised action recognition to gait recognition. Self-supervised approaches such as MS$^2$L \cite{Lin_2020} and Pace Prediction \cite{wang2020selfsupervised} were also chosen for comparison, along with a popular method for unsupervised pretraining in the field of skeleton-based action recognition, Predict \& Cluster \cite{su2019predict}. This latter method uses a sequence-to-sequence LSTM network with fixed decoder to learn discriminative representations. We followed the authors' implementation and pretrain on UWG. The results for direct transfer learning (without fine-tuning) on CASIA-B and FVG are presented in Table \ref{tab:unsupervised-casia} and Table \ref{tab:unsupervised-fvg}, and show that WildGait outperforms existing approaches by a large margin. Our results show that, in the case of gait recognition, the information captured in tracked skeleton sequences of walking people is sufficient for a strong supervisory signal, while plain unsupervised or self-supervised approaches are unsuitable.

Finally, we compared with state-of-the-art skeleton-based gait recognition methods on CASIA-B, with the results presented in Table \ref{tab:casia-sota}. We fine-tuned our network using all the available training data: 62 subjects, all viewpoints and runs. We achieve state-of-the-art results in skeleton-based gait recognition on normal walking (NM) and carry bag variations (CB) by a significant margin. When handling clothing (CL) variation, our method achieves good results relative to other methods, but clothing variation remains a challenging problem for gait-based person identification using skeletons, as heavy clothing significantly affects the manner of walking and also makes certain joints less visible to the pose estimation model. Moreover, UWG, by design, does not contain walking sequences of the same subject with different confounding factors (such as clothing variation). As such, complete disentanglement is cumbersome in our setting. This is noticeable in Figure \ref{fig:fine-tune}, in the case of FVG, where the pretrained model is negatively affected on the clothing variation through the addition of more data, while other variations are more stable.

Our state-of-the-art results are attributed to the large pretraining dataset and the diverse augmentation procedures we employ to make the model invariant to different walking variations and camera viewpoints.

\begin{figure}[hbt!]
    \centering
    \includegraphics[width=1.0\linewidth]{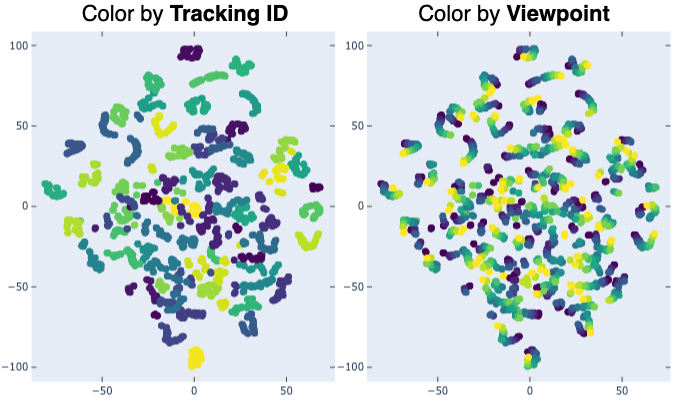}
    \caption{TSNE visualization of the test set from CASIA-B, colored by tracking id (left) and camera viewpoint (right). It is clear that sequences pertaining to the same IDs are close to each other, irrespective of viewpoints.}
    \label{fig:tsne-colors}
\end{figure}

Further, to better understand the behaviour of our model, we make a visualization of embeddings from CASIA-B. Figure \ref{fig:tsne-colors} shows the t-SNE \cite{vanDerMaaten2008} visualization of the test set from CASIA-B. t-SNE has stood the test of time with regards to the visualization power of high dimensional datasets, and can give us important information towards the model performance. As is the case in the numerical evaluation, of interest is the network ability to generalize across viewpoints. We color the same plot two different ways: color by tracking ID - each color is a different person - and by camera viewpoint - each color is a different viewpoint. It is evident that the model clearly clusters walking sequences pertaining to the same subject, regardless of camera viewpoint.

\section{Conclusions}
This work presents a novel weakly-supervised framework, WildGait, for learning informative gait representations from unconstrained environments, without direct human supervision. We show that we can leverage large amounts of video data, surveillance streams, by automatically annotating, filtering and processing walking people to learn discriminative embeddings that generalize well to new individuals, with good disentanglement of confounding factors. We leverage state-of-the-art pose estimation and pose tracking methods to construct Uncooperative Wild Gait (UWG), a large dataset of anonymized skeleton sequences. As far as we know, we are among the first to study pretraining in the context of gait recognition from raw video. Through pretraining on UWG and fine-tuning on downstream recognition tasks, we achieve state-of-the-art results in skeleton-based gait recognition on the CASIA-B benchmarking dataset.

The accuracy of pose-based gait recognition methods is highly dependent on the quality of extracted poses. Large and accurate pose-estimation models come with heavy computational burdens on the processing pipeline, especially in crowded scenes. This suggest a trade-off between accuracy and computational demand / inference time of model-based approaches. We are partly addressing this limitation by releasing the UWG dataset for the pretraining stage, containing over 38K extracted walking skeleton sequences. 
To raise the accuracy for clothing variation scenarios, we aim to further improve our data collection pipeline to include information regarding different clothing for the same person. This implies collecting the same scene (e.g. an office buildings entrance hallway) over a long period of time (ideally a year to include clothing changes due to seasonal variation), and combining information from both facial identification (e.g. FaceNet \cite{DBLP:journals/corr/SchroffKP15}) and person attribute identification models (i.e. HydraNet \cite{DBLP:journals/corr/abs-1709-09930}) for a richer set of automatic annotations. 

One of the main concerns in regards to the broader impact of biometrics-based human identification is privacy. Making use of skeletons for gait recognition softens this concern by relying solely on motion information of people walking, and no identifiable appearance-based information. Furthermore, pose estimation approaches are constantly advancing in terms of performance and efficiency, aiming for real-time inference with negligible to no accuracy loss and requiring less computational resources. This enables pushing skeleton-extraction computation to edge devices, removing the need to upload videos for cloud processing.

\bibliographystyle{named}
\bibliography{ijcai21}

\end{document}